\documentclass{article}

\PassOptionsToPackage{numbers, compress}{natbib}
\usepackage[preprint]{neurips_2020}

\usepackage[utf8]{inputenc} % allow utf-8 input
\usepackage[T1]{fontenc}    % use 8-bit T1 fonts
\usepackage{hyperref}       % hyperlinks
\usepackage{url}            % simple URL typesetting
\usepackage{booktabs}       % professional-quality tables
\usepackage{amsfonts}       % blackboard math symbols
\usepackage{nicefrac}       % compact symbols for 1/2, etc.
\usepackage{microtype}      % microtypography
\usepackage{xcolor}      
\usepackage{amsmath}
\usepackage{amsbsy}
\usepackage{multirow}
\usepackage[pdftex]{graphicx} 
\usepackage{wrapfig} 
\usepackage{footnote}
\makesavenoteenv{tabular}
\makesavenoteenv{table}
\usepackage{subcaption}
\usepackage{algorithm}
\usepackage{algorithmic}

\usepackage{colortbl}
\usepackage{pgfplots}
\usepackage{pgfplotstable}
\usetikzlibrary{pgfplots.colorbrewer}
\pgfplotsset{
    colormap/Greys,
}

\pgfplotstableset{
        /color cells/min/.initial=0,
        /color cells/max/.initial=1000,
        /color cells/textcolor/.initial=,
        color cells/.code={%
            \pgfqkeys{/color cells}{#1}%
            \pgfkeysalso{%
                postproc cell content/.code={%
                    \begingroup
                    \pgfkeysgetvalue{/pgfplots/table/@preprocessed cell content}\value
                    \ifx\value\empty
                        \endgroup
                    \else
                    \pgfmathfloatparsenumber{\value}%
                    \pgfmathfloattofixed{\pgfmathresult}%
                    \let\value=\pgfmathresult
                    \pgfplotscolormapaccess
                        [\pgfkeysvalueof{/color cells/min}:\pgfkeysvalueof{/color cells/max}]
                        {\value}
                        {\pgfkeysvalueof{/pgfplots/colormap name}}%
                    \pgfkeysgetvalue{/pgfplots/table/@cell content}\typesetvalue
                    \pgfkeysgetvalue{/color cells/textcolor}\textcolorvalue
                    \toks0=\expandafter{\typesetvalue}%
                    \xdef\temp{%
                        \noexpand\pgfkeysalso{%
                            @cell content={%
                                \noexpand\cellcolor[rgb]{\pgfmathresult}%
                                \noexpand\definecolor{mapped color}{rgb}{\pgfmathresult}%
                                \ifx\textcolorvalue\empty
                                \else
                                    \noexpand\color{\textcolorvalue}%
                                \fi
                                \the\toks0 %
                            }%
                        }%
                    }%
                    \endgroup
                    \temp
                    \fi
            }%
        }%
    }
}

\usepackage{pifont}
\newcommand{\cmark}{\ding{51}}%
\newcommand{\xmark}{\ding{55}}%

\title{Training With Data Dependent Dynamic \\ Learning Rates}

\author{
 Shreyas Saxena\\
Apple\\
{\tt\small shreyas\_saxena@apple.com}
\and
\textbf{Nidhi Vyas}\\
Apple\\
{\tt\small nidhi\_vyas@apple.com}
\and
\textbf{Dennis DeCoste}\\
Apple\\
{\tt\small ddecoste@apple.com}
}

\begin{document}
\maketitle
\begin{abstract}
Recently many first and second order variants of SGD have been proposed to facilitate training of Deep Neural Networks (DNNs).
A common limitation of these works stem from the fact that they use the same learning rate across all instances present in the dataset.
This setting is widely adopted under the assumption that loss-functions for each instance are similar in nature, and 
hence, a common learning rate can be used.
In this work, we relax this assumption and propose an optimization framework which accounts for difference in loss function characteristics across instances.
More specifically, our optimizer learns a dynamic learning rate for each instance present in the dataset.
Learning a dynamic learning rate for each instance allows our optimization framework to focus on different modes of training data during optimization.
When applied to an image classification task, across different CNN architectures, learning dynamic learning rates leads to consistent gains over standard
optimizers. When applied to a dataset containing corrupt instances, our framework reduces the learning rates on noisy instances, and improves over the
state-of-the-art. Finally, we show that our optimization framework can be used for personalization of a machine learning model 
towards a known targeted data distribution.
\end{abstract}

\section{Introduction}

\begin{wrapfigure}{r}{5cm}
%\begin{figure}
%\centering
\includegraphics[width=\linewidth]{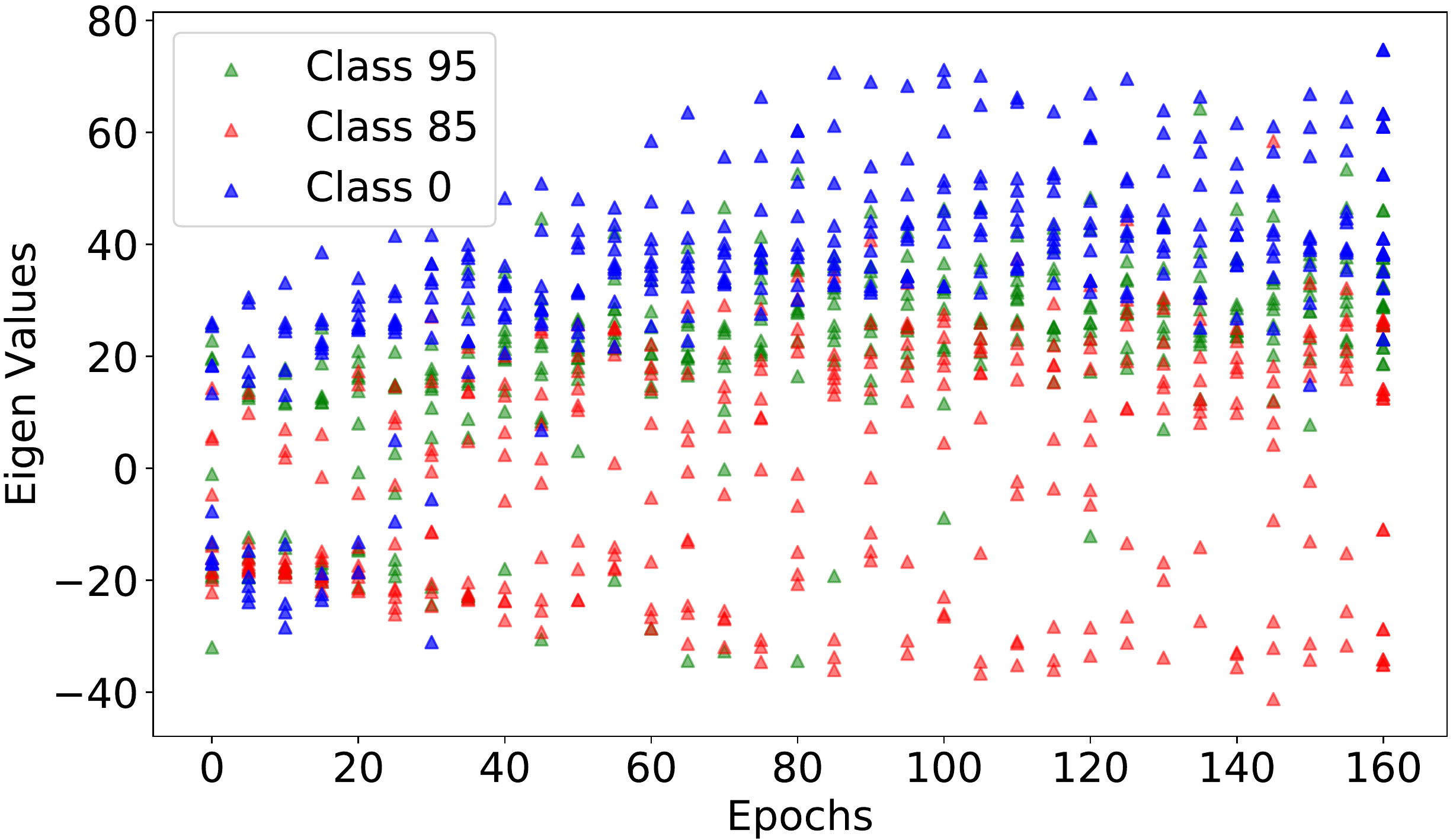}
%\includegraphics[width=0.3\linewidth]{./figures/sum_functions.png}
%\begin{tikzpicture}
%%\draw[very thin,color=gray] (-4.1,-4.1) grid (4.1,4.1);
%\draw[->] (-4.2,0) -- (4.2,0) node[right] {$\theta$};
%\draw[-] (0, 0) -- (0,4.2) node[above] {$f(x)$};
%\draw[thick,green,domain=-3:3,smooth] plot (\x,{0.5*(\x)^2});
%
%\draw[smooth,samples=200,color=blue] plot function{(\cA)* (cos((\cC)*x+(\cD))) + \cB} 
%        node[right] {$f(x) = \cA{} . cos(\cC{} . x + \cD{}) + \cB{}$};
%
%\end{tikzpicture}
%\caption{Toy example motivating the need for learning the learning rate per-data point. We could also show supervised learning, transfer learning,
%semi-supvervised learning, class-imbalance learning as setting the learning rate and schedule for different data-points.}
\caption{Eigen-value spectrum of Hessian for ResNet-18 trained on CIFAR100. 
For three classes, we plot top-10 eigen-values over the course of training. 
The loss function characteristics are different for each class.
%Over the course of optimization, the loss function characetristics are different for each class.
%(Right): Toy figure depicting loss functions for three different instances in a dataset. Length of the red arrow denotes the optimal learning rate.
%As seen from figure, depending upon the characetristics of loss function, the optimal learning rate for each instance is different.}
}
\label{fig:introduction_fig}
%\end{figure}
\vspace{-0.2in}
\end{wrapfigure}

Owing to the non-convex nature of most loss functions, coupled with poorly understood learning dynamics, optimization of Deep Neural Networks (DNNs) 
is a challenging task. Recent work can be broadly divided in two categories: (1) adaptive first order variants of SGD, such as ADAMw
\cite{loshchilov2018decoupled}, RMSprop \cite{hinton2012neural} and Adam \cite{kingma2014adam},
and (2) second order methods such as K-FAC \cite{martens2015optimizing}. 
These methods adapt the optimization process per parameter and obtain improved convergence and
generalization. The objective function for these frameworks is to minimize finite-sum problems:
$\mathcal{L}(\theta)  = \frac{1}{N} \sum_i^{N} L^i(\theta)$,
%\begin{equation*}
%\begin{split}
%\theta^{t+1} = \theta^t - \frac{\lambda}{N} \sum_i^N (\hat{\lambda}^{i,t} \cdot \frac{\partial L^i}{\partial \theta^t})
%\end{split}
%\end{equation*}
%such as those encoutered in empirical risk minimization,
where $L^i(\theta)$ denotes the loss on a single training data point. 
Standard optimization frameworks assume that the $N$ loss functions come from the same distribution, 
and share characteristics such as the curvature, Lipshitz constant, etc. 
However, in practice, this is not true. Figure \ref{fig:introduction_fig} highlights
the difference in loss function characteristics over the course of training for classes present in CIFAR100.

In contrast to prior work, which models the learning dynamics of \textit{sum} of finite problems,
in this work, we account for differences in characteristics of underlying loss functions in the sum. %by modelling the learning dynamics per instance. 
%In this work, we relax this assumption and account for difference in characteristics of the loss functions.
%Instead of modelling the learning dynamics for each model parameter, 
%we model learning dynamics per instance of the dataset. 
More specifically, for each instance in the dataset, we associate a dynamic learning rate, and learn it along with model parameters.
The SGD update rule for our framework takes the following form:
\begin{equation*}
\theta^{t+1} = \theta^t - \frac{\lambda}{N} \sum_i^N (\mathbf{w}^{i,t}_{inst} \cdot \frac{\partial L^i}{\partial \theta^t})
\end{equation*}
where $L^i$, $\lambda$, $\mathbf{w}^{i,t}_{inst}$ denotes loss, model parameter learning rate, and multiplicative learning rate correction for 
instance $i$ at time step $t$. 
Having a dynamic learning rate per instance allows our optimization framework to focus on different modes of data during the course of optimization.
%Unlike existing optimization frameworks, having a dynamic learning rate per instance allows our optimization framework to 
%account for differences in loss function characteristics, to focus on different modes of data during the 
%course of learning. 
Instead of using a heuristic, we learn the learning rate for each instance via meta-learning using a held out meta set. 
We also extend our framework to learn a dynamic learning rate per class. 
%We show that our 
%optimization frameworks reduces variance in the optimization process, is robust to annotation noise in datasets 
%and achieves better generalization and convergence compared to widely used optimizers.

The main contributions of our work are:
\begin{enumerate}
\item We present an optimization framework which learns an adaptive learning rate per instance in the dataset to account 
for differences in the loss function characteristics across instances. We extend this framework to learn adaptive learning rates per class. 
  We also show that learning dynamic weight-decay facilitates learning  dynamic learning rates on data.
%We further generalize our framework to learning the learning rate per class, and dataset. 
%\item To the best of our knowledge, we are the first work to show gains by learning dyanmic weight regularization. 
%More importantly, we also show that learning a dynamic weight decay regularization is important when one would like to learn dynamic learning rate on data.
%\item We show that learning class-level learning rates, allows us to model class-imbalance problem.
%\item We show that by learning the learning rate, we can automate the transfer learning process and obtain state of the art results for the same.
\item We show that learning adaptive learning rate on data leads to variance reduction, and explains faster convergence and improved accuracy.
\item We show that in presence of noisy data, our framework reduces the learning rates on noisy instances, and prioritizes learning from clean instances. 
Doing so, we our method outperforms state-of-the-art by a significant margin.
\item We show that our framework can be used for personalization of machine learning models towards a targeted distribution.

%We show that by learning the learning rate for samples in the dataset, we can personalize the DNN model towards a targeted distribution. 
%\item We show that learning a dynamic learning rate over data points leads to variance reduction in the gradient estimator, and explains faster convergence and
%improved accuracy.
%\item We show that learning the learning rate per data-point changes the course of optimization by reducing the top-eigenevalues of hessian.
%\item We show that learning the learning rate per data-point and model-parameter (eg. ADAM) accelerates convergence (CIFAR100 vs NLP).
%\item We show that learning the learning rate per data-point improves semi-supervised learning.
%\item We show that learning the learning rate per data-point prevents memorization of random-labels.
%\item We show that keeping a history of instance-level weights is important, instead of online optimization.
%\item We show that dot-product of the gradient with the optimal direction ($\theta-theta_{opt}$) is high for our optimizer.
%\item We show that we can learning instance level learning rates for regression formulation, and train better auto-encoder models.

\end{enumerate}

\newcommand{\cA}{1}%    Cste . fct
\newcommand{\cB}{0}%    Cste + fct
\newcommand{\cC}{1}%    Cste . var
\newcommand{\cD}{0}%

\section{Learning the learning rate for data}
As mentioned earlier, the main goal of our work is to learn a dynamic learning rate for instances present in the train dataset. 
In this section, we first formalize this intution and present the framework for learning instance level learning rates. 
Next, we will show how our framework can be extended to learn class level learning rates.
We derive our method using stochastic gradient descent (SGD) as the optimizer for model paramters, however,
extension to other class of optimizers can be done in a similar manner.

\subsection{Learning instance level learning rate}
\label{sec:dynamic_instance_weighting}
Let $\left\{\left(\mathbf{x}^i_{train}, y^i_{train}\right)\right\}^N_{i=1}$ denote train set, 
where $\mathbf{x}^i \in \mathbb{R}^d$ denotes a single data point in the train set and $y^i \in \{1, ..., k\}$ denotes 
the corresponding target. Let $f(\mathbf{x}, \theta^t)$ and $\theta^t$ denote the model and  model's 
parameters at step $t$ respectively. Let  $L(\mathbf{x}^i_{train}, y^i_{train}; \theta^t)$ denote an arbitrary
differentiable loss function on data point $i$ at time step $t$. 
In what follows, we denote $L(\mathbf{x}^i_{train}, y^i_{train}; \theta^t)$ as $L_{train}^{i}(\theta^t)$.
%Note, learning the learning rate for each instance is equivalent to learning weighting for each instance.
Let $\mathbf{w}_{inst}^t \in \mathbb{R}^N$ denote the instance level parameters at time step $t$. 
Instance level parameters weigh contribution of instances in the gradient update (see Equation \ref{eq:sgd_update}), and can be interpreted as a
%Note, since instance level parameters perform a linear weighting, they can be interpreted as a 
multiplicative learning rate correction over the learning rate of model parameter's optimizer. Note, learning the learning rate for each instance
is equivalent to learning weighting for each instance.
For the ease of explanation, we choose the latter formulation.
%$\mathcal{L}_{train}(\theta^t, \mathbf{w}_{inst}^{t})$ denotes the weighted loss on the train set. 
Our goal in optimization is to solve for optimal $\hat{\theta}$ by minimizing the weighted loss on the train set:
\begin{align}
\mathcal{L}_{train}(\theta^t, \mathbf{w}_{inst}^{t}) &= \frac{1}{N} \sum_{i=1}^N w^{i, t}_{inst} \cdot L_{train}^{i}(\theta^t)
\label{eq:sgd_update} \\
\hat{\theta}(\mathbf{w}_{inst}^{t}) &= \arg \min_{\theta} \mathcal{L}_{train}(\theta, \mathbf{w}_{inst}^{t})
\end{align}
The solution to above equation is a function of $\mathbf{w}_{inst}^{t}$, which is not known a priori.
Setting $\mathbf{w}_{inst}^t$ as $\mathbf{1}$ at all time steps recovers the standard gradient descent optimization framework, but that might not be optimal.
This brings us to the question: \textit{What is the optimal value of $\mathbf{w}^{t}_{inst}$ 
\textit{i.e.} the learning rate for data points at time step $t$?}
%Solving for $\mathbf{w}_{inst}$ to minimize the train loss $\mathcal{L}_{train}$ leads to a degenerate solution $\mathbf{w}_{inst} = \mathbf{0}$.
 
\paragraph{Learning dynamic instance level learning rate via meta-learning}
In contrast to model parameters, whose optimal value is approximated by minimizing the loss on train set, we can not approximate the optimal value of
$\mathbf{w}^{t}_{inst}$ by minimizing the loss on train set. Doing so, leads to a degenerate solution, where $\mathbf{w}_{inst}^{t} = \mathbf{0}$.
In principle, the optimal value of $\mathbf{w}^{t}_{inst}$ is the one, which when used to compute the gradient update at time step $t$, minimizes the error 
on a held-out set (referred as meta set) at convergence, \textit{i.e} $\hat{\mathbf{w}}_{inst}^{t} = 
\arg \min_{\mathbf{w}_{inst}^{t}} \mathcal{L}_{meta}(\hat{\theta}(\mathbf{w}_{inst}^{t}))$. Here $\hat{\theta}$ denotes model parameters at convergence. 
The sequence of model updates from time step $t$ till convergence ($\theta^t \rightarrow \hat{\theta}$) can be written as a feed forward computational graph, 
allowing us to backpropagate meta-gradient (gradient on meta set) to $\mathbf{w}^{t}_{inst}$.  However, this is not feasible in practice due to: 
(1) heavy compute for backpropagating through time steps, (2) heavy memory foot-print from saving all intermediate representations and 
(3) vanishing gradient due to backpropagation through time steps. 
%There are some recent works whic

To alleviate this issue, we approximate the meta-gradient at 
convergence with the meta-gradient at time step $t+1$. %\textit{Should we speak why this approximation makes sense?}
%More formally, let $\left\{\left(\mathbf{x}^j_{meta}, y^j_{meta}\right)\right\}^M_{j=1}$ denote the meta set, which unless stated otherwise comes from the same 
%distribution as train and test set. 
More formally, we sample a mini-batch from train set, and write one step SGD update on 
model parameters (${\theta^{t+1}}$) as a function of instance parameters at time step 
$t$ (see equation \ref{eq:one_step_sgd_roll_out_instance}). The one step update is used to compute loss 
on meta set $\mathcal{L}_{meta}({\theta^{t+1}})$, which is then used to compute the meta-gradient on instance parameters:
%The stochastic gradient on an instance parameter is given by:
\begin{align}
%\mathcal{L}_{meta}(\theta^{t+1}) &= \frac{1}{M} \sum_{j=1}^M L_{meta}^j (\theta^{t+1}).\\
{\theta^{t+1}} (\mathbf{w}_{inst}^{t}) &= 
\theta^{t} - \frac{\lambda}{N} \sum_{i=1}^N \mathbf{w}^{i,t}_{inst}\frac{\partial L_{train}^i(\theta^t)}{\partial \theta^{t}}
\label{eq:one_step_sgd_roll_out_instance} \\
\frac{\partial \mathcal{L}_{meta}({\theta^{t+1}})}{\partial w^{i,t}_{inst}} &= 
\frac{\partial \mathcal{L}_{meta}({\theta^{t+1}})}{\partial {\theta^{t+1}}} 
\cdot \frac{\partial {\theta^{t+1}}}{\partial w^{i,t}_{inst}} 
%\nonumber 
\\
\frac{\partial \mathcal{L}_{meta}({\theta^{t+1}})}{\partial w^{i,t}_{inst}} &=  
\frac{-\lambda}{N} \cdot \frac{\partial \mathcal{L}_{meta}({\theta^{t+1}})} {\partial {\theta^{t+1}}} \cdot 
\big[ {\frac{\partial L_{train}^i}{\partial \theta^{t}}} \big]^T \label{eq:meta_gradient_instance}
\end{align}
%Here, in equation \ref{eq:meta_gradient_instance}, we make an approximation $\frac{\partial \theta_t}{\partial w^{i,t}_{inst}} \approx 0$.
Here, $\lambda$ and $N$ corresponds to the learning rate of the model optimizer and number of samples in train mini-batch respectively.
Using the meta-gradient on instance-parameters we update the instance-parameters using first order gradient update rule 
(see equation \ref{eq:t_plus_1_update_inst_params}).
%The updated value of instance-parameters is then used to update the model parameters (see equation \ref{eq:t_plus_1_update_model}).
\begin{align}
w^{i,t+1}_{inst} &= w^{i,t}_{inst} - \lambda_{w_{inst}} \frac{\partial \mathcal{L}_{meta}({\theta^{t+1}}^*)}{\partial w^{i,t}_{inst}}
\label{eq:t_plus_1_update_inst_params} 
%\theta^{t+1} &= 
%\theta^{t} - \frac{\lambda}{n} \sum_{i=1}^n \mathbf{w}^{i,t+1}_{inst}\frac{\partial L_{train}^i(\theta^t)}{\partial \theta^{t}}
%\label{eq:t_plus_1_update_model}
\end{align}
The pseduo code for our method is outlined in Algorithm \ref{alg:example}.

\begin{algorithm}[h]
\vspace{0mm}
\renewcommand{\algorithmicrequire}{\textbf{Input:}}
\renewcommand{\algorithmicensure}{\textbf{Output:}}
\caption{Learning algorithm for learning the learning rate per instance}
\label{alg:example}
\begin{algorithmic}[1]  \small
	\REQUIRE Train set $\mathcal{D}_{train}$, meta set $\mathcal{D}_{meta}$, model learning rate $\lambda$, 
        instance parameter learning rate
        $\lambda_{inst}$, max iterations $T$.
	\ENSURE  Model parameters at convergence $\theta^{T}$.
	%\REPEAT
	\STATE Initialize model parameters $\theta^{0}$ and instance level parameters $\mathbf{w}_{inst}^0$.
	\FOR{$t=0$ {\bfseries to} $T-1$}
	\STATE $\{x_{train}^i,y_{train}^i\} \leftarrow$ SampleMiniBatch($\mathcal{D}_{train}$).
	\STATE $\{x_{meta}^j,y_{meta}^j\} \leftarrow$ SampleMiniBatch($\mathcal{D}_{meta}$).
	\STATE $\{\mathbf{w}_{inst}^{t}\} \leftarrow$ SampleInstanceParameters.
	\STATE Update model parameters, and express $\theta^{t+1}$ as a function of $\mathbf{w}_{inst}^{t}$ by Eq. (\ref{eq:one_step_sgd_roll_out_instance}).
	\STATE Update $\mathbf{w}_{inst}^{t+1}$ by Eq. (\ref{eq:t_plus_1_update_inst_params}).
%	\STATE Update $\theta^{t+1}$ by Eq. (\ref{eq:t_plus_1_update_model}).
	\ENDFOR
	%\UNTIL{$noChange$ is $true$} 
\end{algorithmic}
\end{algorithm}
%\vspace{2mm}

\paragraph{Analysis of meta-gradient on instance level parameters}
From equation \ref{eq:meta_gradient_instance}, we can observe that the meta-gradient on instance parameter 
$\mathbf{w}_{inst}^{i, t}$ is proportional to the dot product of $i^{th}$ training sample's gradient on model 
parameters at time step $t$ with meta-gradient on model parameters of samples in the meta set 
at time step $t+1$. Therefore, training samples whose gradient aligns with the gradients on meta-set 
will obtain a higher weight, leading to an increased learning rate. 
The converse holds true as well. For example, if an instance in train set has wrong label, its gradient will not align with gradient of clean instances in the
meta set. Over the course of learning, the corrupt instance will end up obtaining a lower value of instance parameter.

\subsection{Learning class level learning rate}
%In the previous section, we have detailed how we can use our framework to learn learning rate per instance present in the dataset.
While instance parameters have the flexibility to adapt to each instance present in the dataset, 
number of parameters grow with the size of dataset. To alleviate this issue, another way we can partition a dataset is 
by leveraging the class membership of data points.
More specifically, we can learn a learning rate for each class, shared by all the instances present within the class.
Let $\mathbf{w}_{class}^t \in \mathbb{R}^N$ denote the class parameters at time step $t$. 
Similar to equation (\ref{eq:one_step_sgd_roll_out_instance}), we can write the one step look ahead update 
as a function of class parameters, and use it compute the meta-gradient on the class parameters:
\begin{align}
%{\theta^{t+1}}^* (w_{class}^{y_i, t}) &= 
%\theta^{t} - \frac{\lambda}{n} \sum_{i=1}^n w^{y_i,t}_{class}\frac{\partial L_{train}^i(\theta^t)}{\partial \theta^{t}} \\
\frac{\partial \mathcal{L}_{meta}({\theta_{t+1}}^*)}{\partial \mathbf{w}^{c, t}_{class}} &=  
                            -\frac{\lambda.n_c}{N} \cdot
                            \frac{\partial \mathcal{L}_{meta}({\theta^{t+1}}^*)}{\partial {\theta_{t+1}}^*}  \cdot 
                            \big[\frac{1}{n_c} {\sum\limits_{\substack{i \\ y_i = c}}^{n_c} \frac{\partial L_{train}^i}{\partial
                                                \theta_{t}}} \big]^T 
                                                \label{eq:meta_gradient_class}
\end{align}
Here, $w^{c, t}_{class}$ denotes the weight for $c$ (target class for $i^{th}$ train data point), and 
$n_c$ denotes number of samples in class $c$. 
As seen in equation \ref{eq:meta_gradient_class}, the meta-gradient on class parameters is 
proportional to the dot-product of meta-gradient on model parameters with gradient on model parameters from train set, averaged over
instances belonging to class $c$. We provide detailed derivation in supplementary material.

\subsection{Meta-learning weight decay regularization}
Use of weight-decay as a regularizer is a de-facto standard in training Deep Neural Networks (DNNs). 
However, the exact role weight-decay plays in optimization of modern DNNs is not well understood 
\cite{golatkar2019time, hoffer2018norm, loshchilov2018decoupled, zhang2018three}. In this section, we highlight the importance
of learning the weight-decay coefficient along with the learning rate for dataset, class or instances. 
For ease of explanation, let us consider the case where we are interested in learning instance level learning rate, 
and we have the standard weight-decay term added as a regularizer.
\begin{align}
\mathcal{L}_{train}(\theta^t, w_{dataset}^{t}) &= \frac{1}{N} \sum_{i=1}^N \mathbf{w}^{i,t}_{inst} \cdot L_{train}^{i}(\theta^t) + 
\lambda_{wd} \|\theta^t\|
\end{align}
Here $\lambda_{wd}$ denotes the weight-decay coefficient. During the course of optimization, 
regardless of the magnitude of the first term, the contribution of the second term in gradient update is fixed.
In our experiments, we found this to be problematic. 
When meta-gradient reduces the magnitude of instance parameter $\mathbf{w}^{i,t}_{inst}$, it leads to a 
relative increase of weight-decay component in the gradient update. This leads to destablization of the training.
We solved this problem by treating the weight-decay coefficient as a learnable parameter, which is learnt along with the learning rate of class and instances
using meta learning setup.

\section{Experiments}

\subsection{Implementation details}
\label{subsec:implementation_details}
Unless stated otherwise, the following implementation details hold true for all experiments in the paper. 
We use SGD optimizer (without momentum and weight-decay) to learn instance and class level learning rates.
We use same batch-size to sample batches from train-set and meta-set. 
Apart from clamping negative learning-rates to 0, we do not employ any form of 
regularization, and rely on the meta-gradient to regularize the learning process.
We perform $k$-fold cross validation, where the held-out set is used as both: meta-set and validation-set (for picking best configuration).
For reporting final numbers, we average out dynamic learning rate trajectory for each class and instance, and use it to train on the full train-set.
We ensure all methods use the same amount of training data. We report mean and standard-deviation computed over 3 runs.

\subsection{Image Classification on CIFAR100}
In this section, we show efficacy of our optimization framework when applied to the task of image classification on CIFAR100 \cite{krizhevsky2009learning}
dataset. CIFAR100 dataset contains 100 classes, 50,000 images in the train set and 10,000 images in the test set. 
Therefore, in our framework, along with the model parameters, we learn 100 and 50,000 dynamic learning 
rates for class and instances respectively. 
We evaluate our framework with ResNet18 \cite{he2016deep}, VGG16 \cite{simonyan2015very}.
We use standard setup for training both architectures (details in supplementary).

In Figure \ref{fig:comparison_optimizer}, we compare our optimization framework to other optimizers commonly used in the deep learning community.
%We tune all of these optimizers by doing a grid search over their hyper-parameters (details in supplementary section).
Similar to results in \cite{zhang2019lookahead}, when tuned appropriately, apart from ADAM, all other optimizers obtain performance comparable to SGD. In both settings, learning class or instance level learning rate, we outperform these standard optimizers by a significant margin. 
These gains over standard optimizers can be attributed to the fact that 
our framework adapts the optimization process over samples in dataset instead of model-parameters.

Across both architectures, learning instance level learning rates performs more favorably 
compared to learning class level learning rates. This validates our hypothesis: 
loss functions for samples within a class might have different characteristics, and
might benefit from learning sample specific learning rates. 
However, class level parameters get more frequent updates compared to instance level parameters, and hence can achieve faster convergence (see
Figure \ref{fig:comparison_optimizer}, middle).

\begin{figure}
\begin{minipage}{\textwidth}
\begin{minipage}[]{0.3\textwidth}
        \centering
        \resizebox{1\columnwidth}{!}{
          \begin{tabular}{cccc}
            \toprule
                                                     &    ResNet18           &   VGG16               \\    
            \midrule                                              
               SGD                                   & 77.5 $\pm$ 0.0        &   75.4 $\pm$ 0.2      \\        
               Momentum \cite{polyak1964some}        & 77.1 $\pm$ 0.3        &   74.1 $\pm$ 0.2      \\        
               Adam \cite{kingma2014adam}            & 74.4 $\pm$ 0.2        &   72.2 $\pm$ 0.5      \\        
               AdamW \cite{loshchilov2018decoupled}  & 77.4 $\pm$ 0.1        &   74.5 $\pm$ 0.2      \\        
               Polyak \cite{polyak1992acceleration}  & 78.0 $\pm$ 0.3        &   74.5 $\pm$ 0.3      \\        
               LookAhead \cite{zhang2019lookahead}   & 77.2 $\pm$ 0.4        &   75.6 $\pm$ 0.2   \\        
            \midrule                                                                     
               Instance-level                        & \bf{78.6 $\pm$ 0.2}   &   \bf{76.5 $\pm$ 0.1} \\        
               Class-level                           & 78.3 $\pm$ 0.2        &   76.2 $\pm$ 0.2      \\        
        %       Instance + Class                      &      R                &          R            \\        
            \bottomrule
          \end{tabular}
            }

\end{minipage}
\begin{minipage}[]{0.33\textwidth}
        \includegraphics[width=\linewidth]{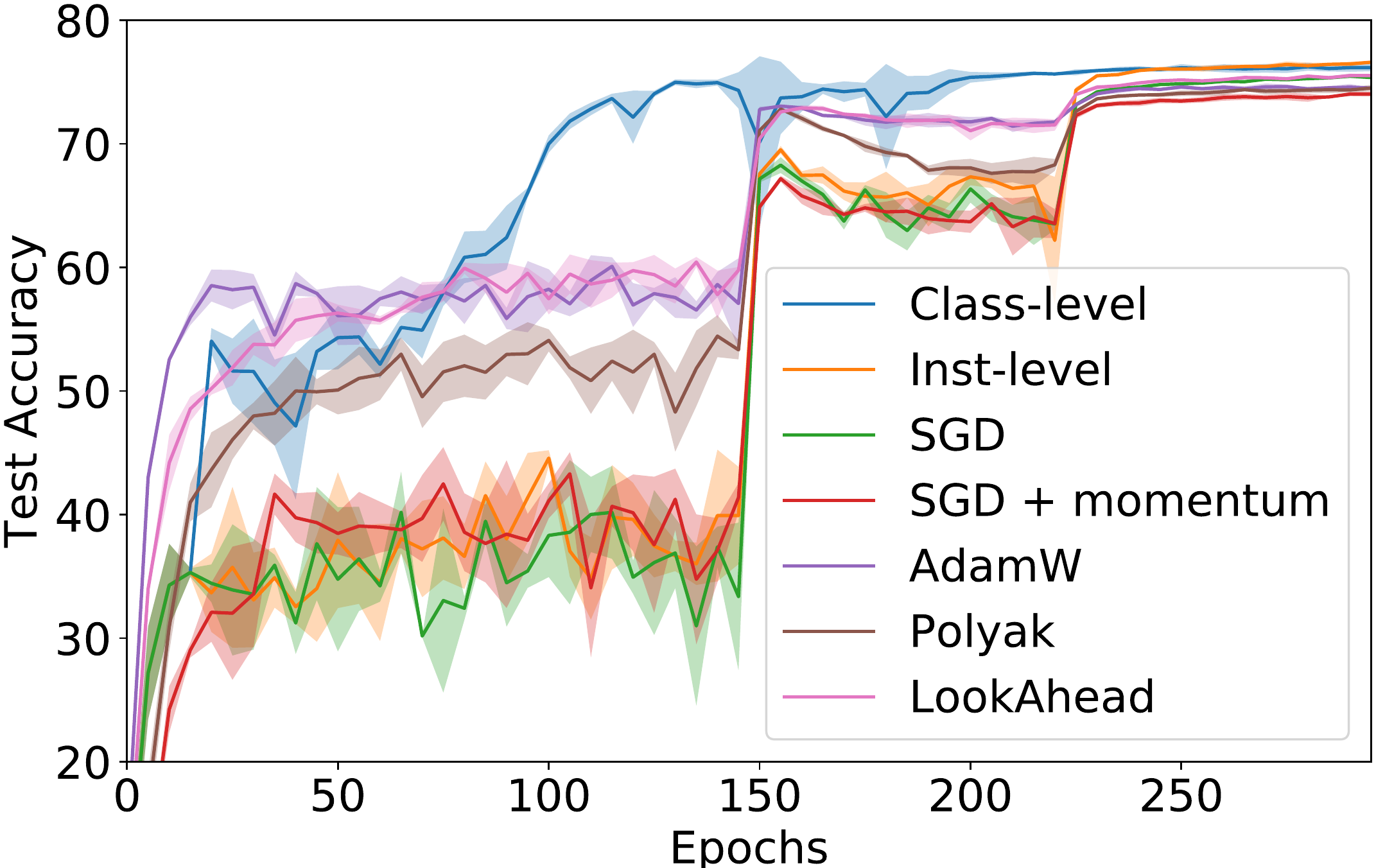}
\end{minipage}
\begin{minipage}[]{0.33\textwidth}
        \includegraphics[width=\linewidth]{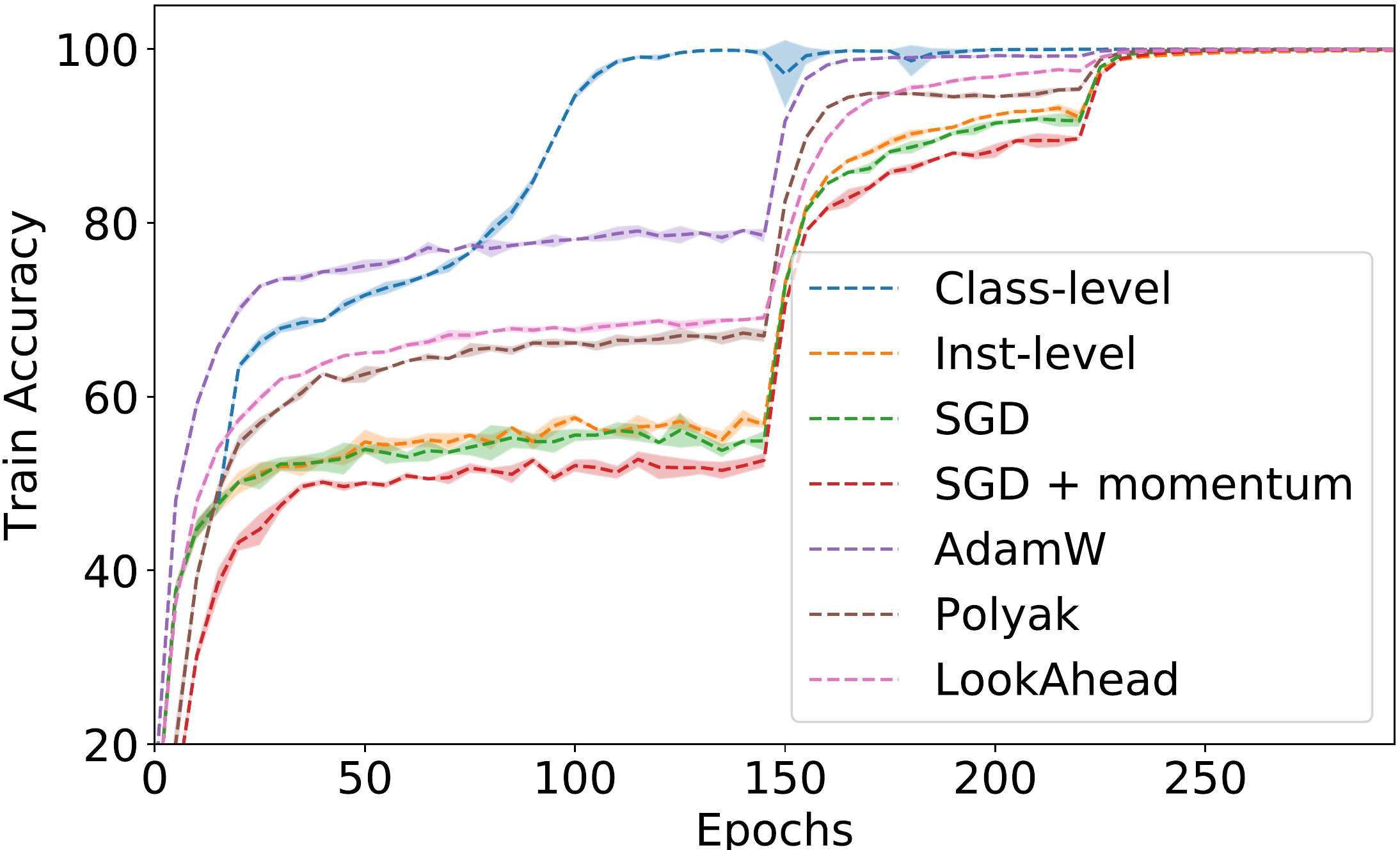}
\end{minipage}
\end{minipage}
\caption{\textbf{Table}: Comparison of class and instance level optimizer with different optimizers on CIFAR100.
We do a grid search on learning rate and weight-decay for other optimizers (see supplementary). 
Lookahead and Polyak are wrapped around SGD.
\textbf{Figures}: Test and train accuracy for different optimizers for VGG16. 
Using our optimizer leads to reduced variance, and better generalization.
}
\label{fig:comparison_optimizer}
\end{figure}

\subsection{Analysis of optimization framework}
In this section, we will analyze different components of our optimization framework.

\begin{figure}
\centering
\includegraphics[width=0.23\linewidth]{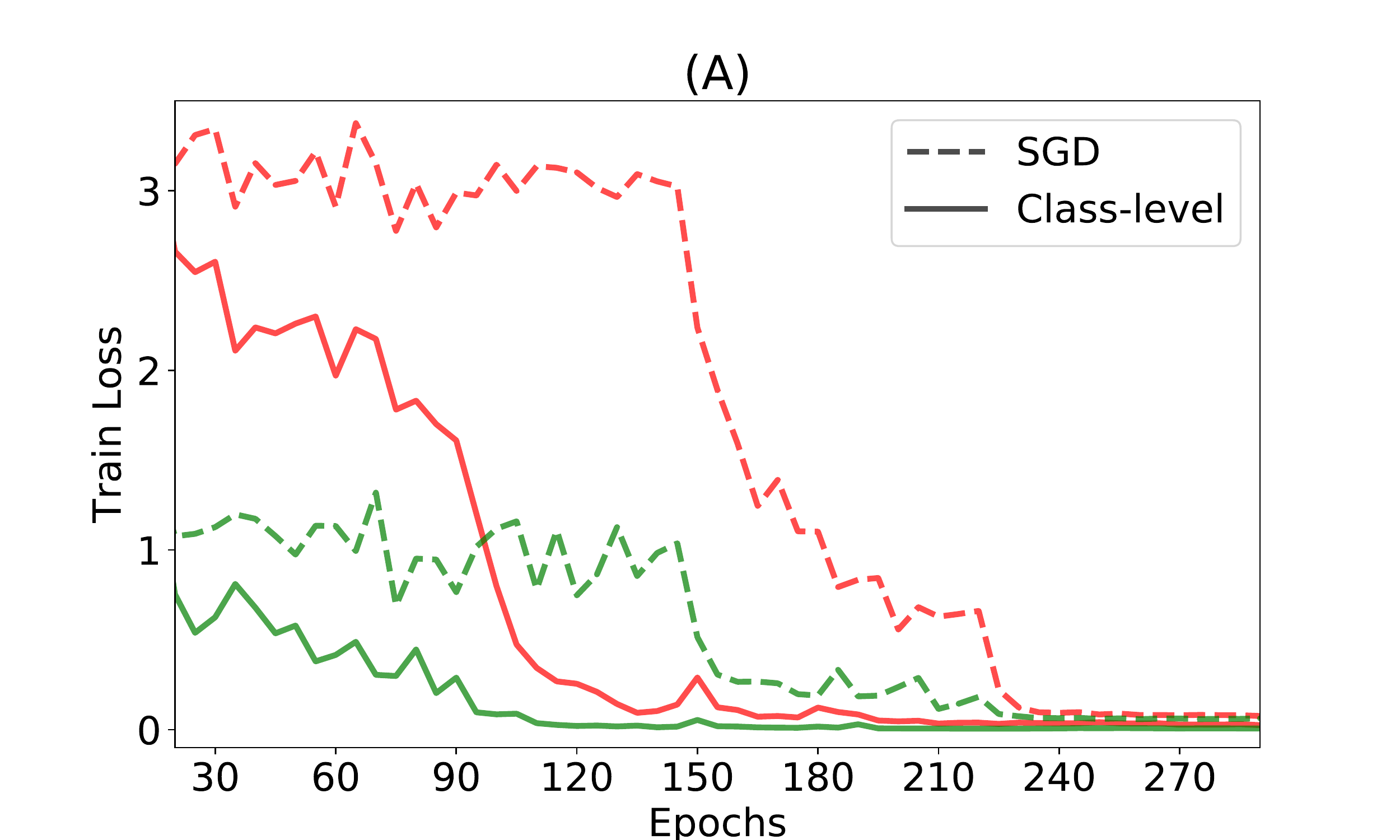}
\includegraphics[width=0.24\linewidth]{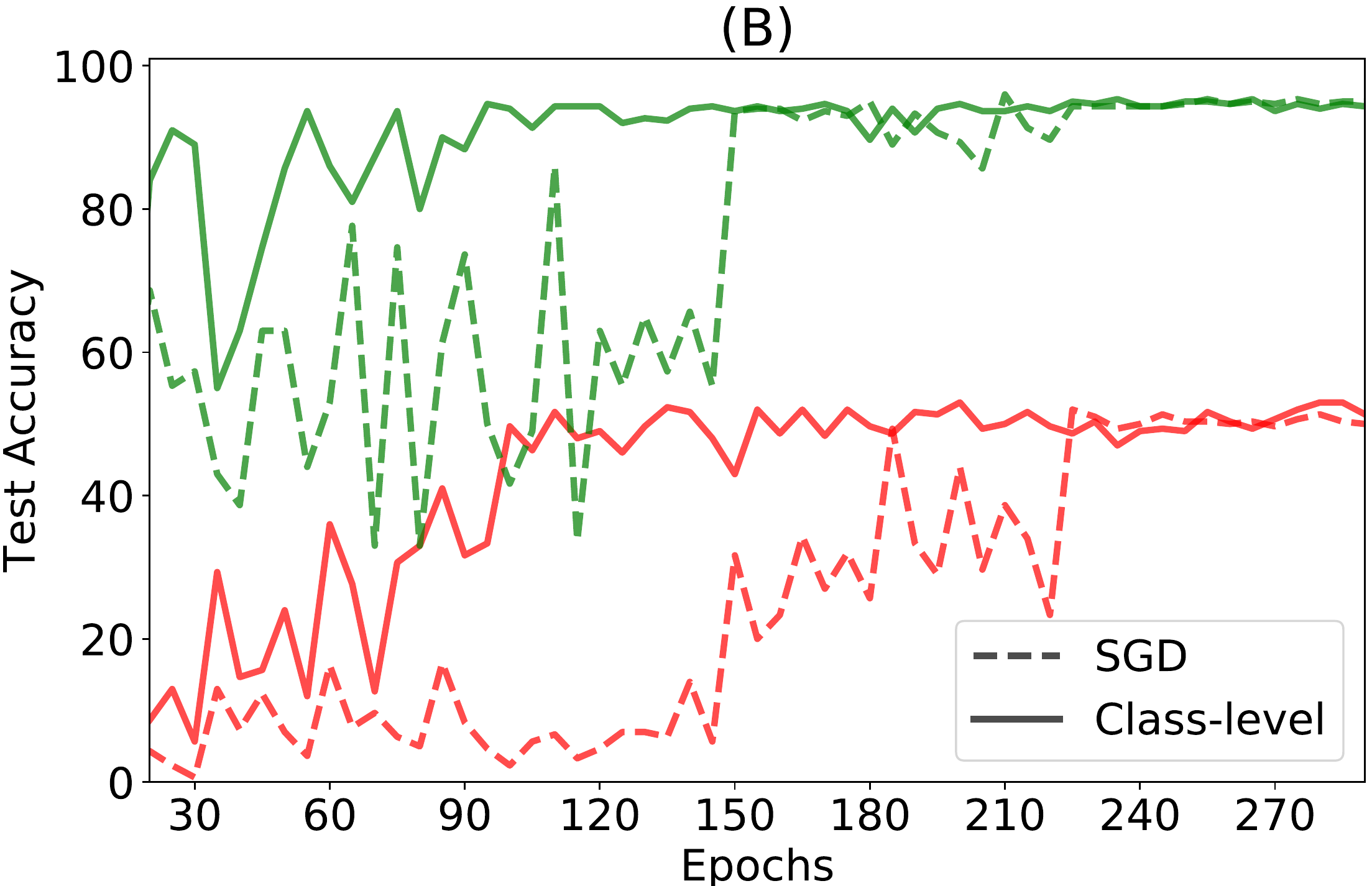}
\includegraphics[width=0.24\linewidth]{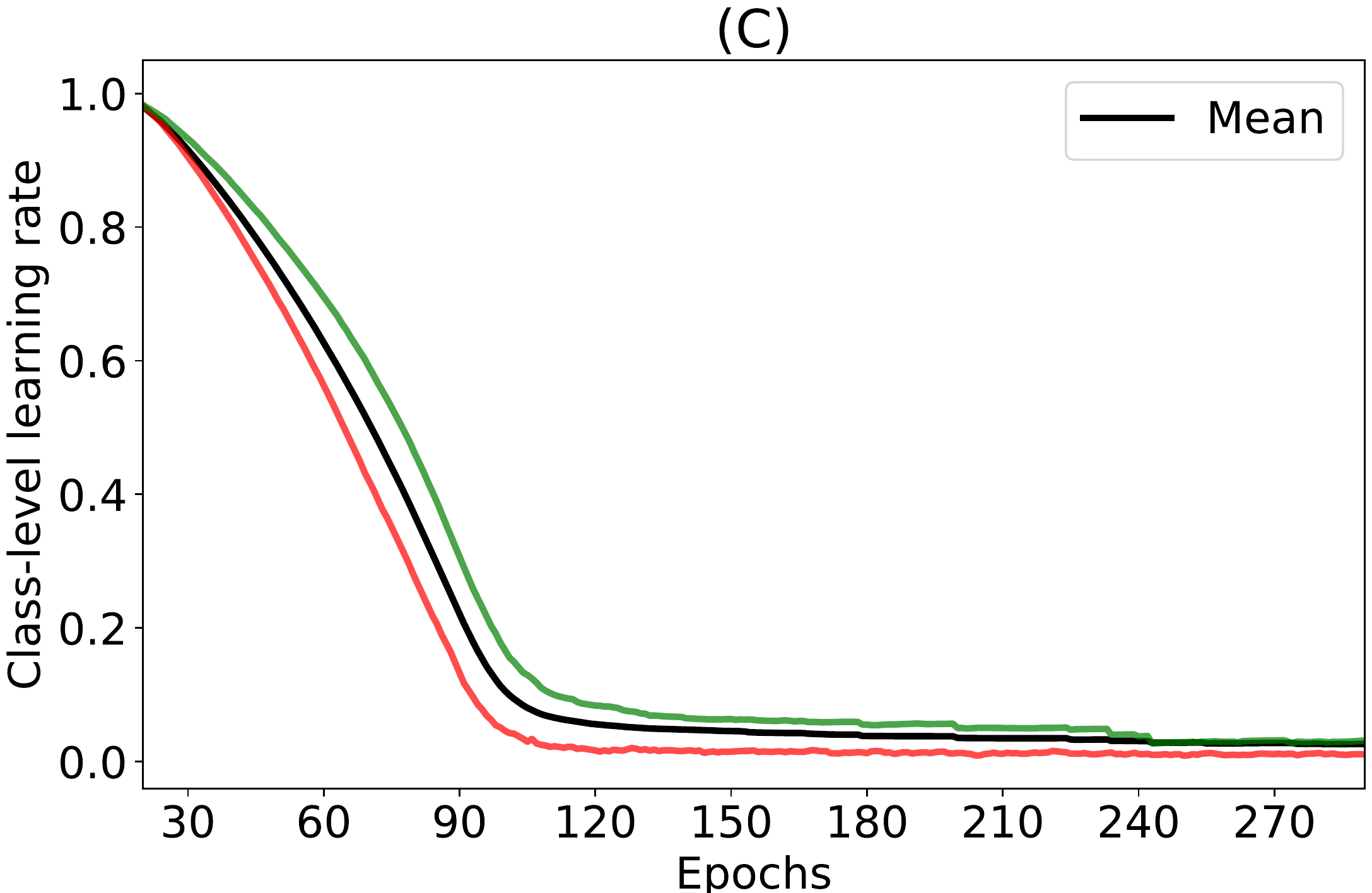}
\includegraphics[width=0.24\linewidth]{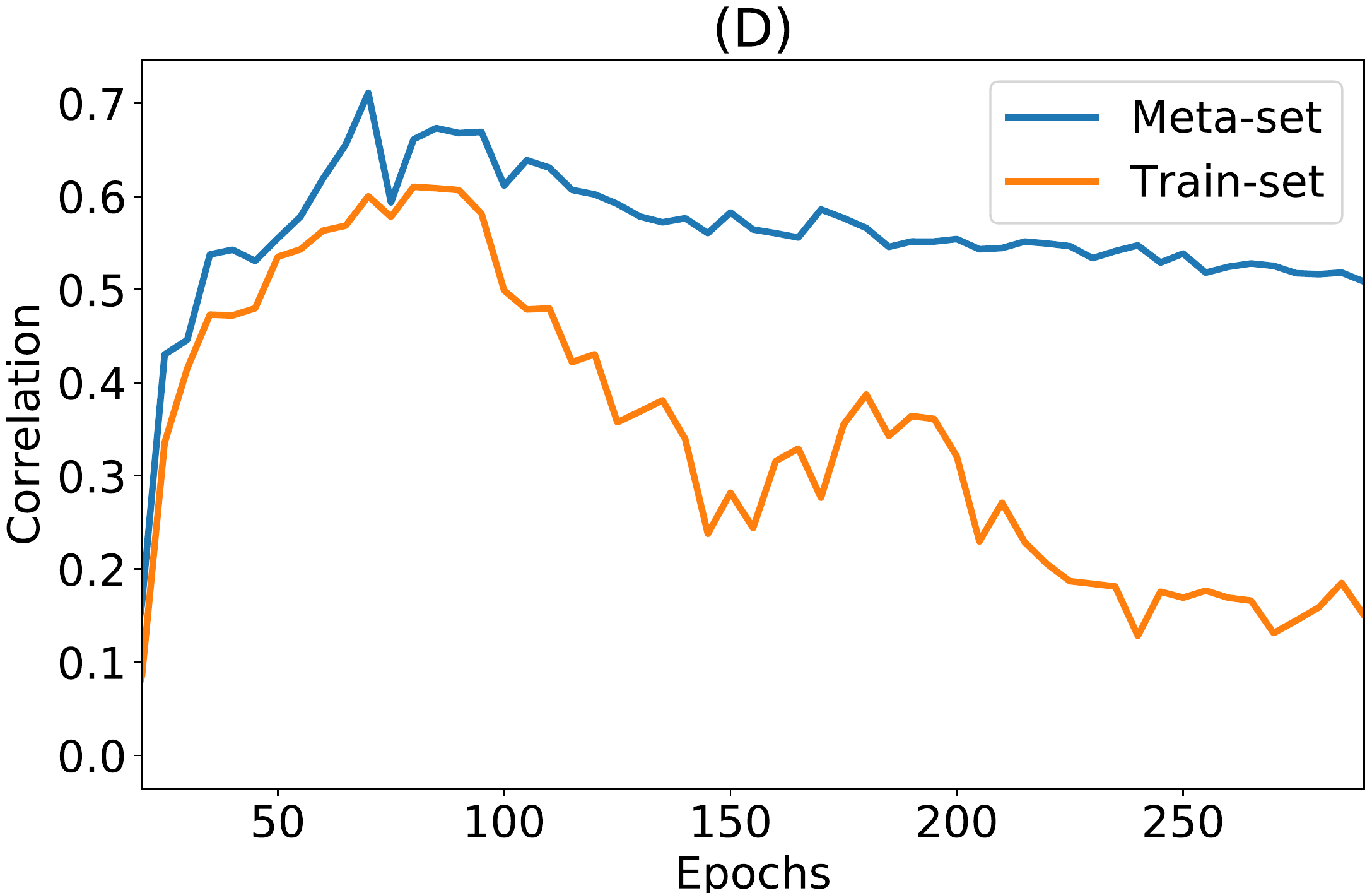}
\caption{To speed up convergence, our framework reduces learning rate on classes with worse performance on meta set (see text for details). Learning
dynamics for class with best (green) and worst (red) performance at convergence. \textbf{A:}Train loss. \textbf{B:} Test accuracy. \textbf{C:} Dynamic
learning rate for the classes, along with  mean learning rate for all classes (black). \textbf{D:} At each epoch, we plot the correlation of learning rate of a
class with their performance on heldout meta-set and train-set.}
\label{fig:importance_learning_rate}
\end{figure}

\paragraph{Variance reduction via dynamic learning rates on data}
One key hypothesis of our work is: loss functions for different data points can have different characteristics, and hence might 
benefit from different learning rates. In Figure \ref{fig:importance_learning_rate} we empirically verify this property on CIFAR100, using class-level learning 
rates for training VGG16. In Figure \ref{fig:importance_learning_rate} (A, B) we plot the train loss and test accuracy for the best (green) and the worst (red) performing class at convergence. 
Shared learning rate used by SGD works well for the green class, but is not able to optimize the red class (until learning rate decay at epoch 150).
In contrast, our method accounts for class performance on the meta-set, and reduces the learning rate for each class in proportion to that (see
Figure \ref{fig:importance_learning_rate}, C and D). This result provides an interesting view to our method from the perspective of variance reduction. 
In contrast to standard setting where methods have been proposed to reduce variance in the gradient estimator for the entire mini-batch, our work performs 
selective variance reduction. Lowering the learning rate for classes with worse performance
will lower the overall variance in gradient estimator, leading to faster and stable convergence. 
In light of recent work \cite{defazio2019ineffectiveness}, which shows ineffectiveness of standard variance reduction framework in deep learning, 
our results indicate that performing selective variance reduction could be an interesting direction to explore.

\paragraph{Importance of history}
\begin{wrapfigure}{r}{5cm}
%\begin{table}[ht]
  \centering
    \resizebox{0.35\columnwidth}{!}{
  \begin{tabular}{ccccc}
    \toprule
                              & History &      ResNet18     &   VGG16          \\
    \midrule
    \multirow{2}{*}{Instance} & \cmark  &   78.6 $\pm$ 0.2  & 76.5 $\pm$ 0.1   \\
                              & \xmark  &   77.8 $\pm$ 0.2  & 75.4 $\pm$ 0.2   \\\midrule
    \multirow{2}{*}{Class}    & \cmark  &   78.3 $\pm$ 0.2  & 76.2 $\pm$ 0.2   \\
                              & \xmark  &   77.8 $\pm$ 0.1  & 75.7 $\pm$ 0.1    \\
    \bottomrule
  \end{tabular}
    }
  \captionof{table}{Impact of retaining history on CIFAR100 for image-classification.}
%  As shown from the table, not retaining the history of learnt learning rates leads to worse performance.}
  \label{tab:impact_of_learnt_history}
% qdjgxt8hwk: ResNet Instance
% nfdcrenhfw: VGG16 Instance
% 3mtu4shu2f: VGG16 Class
% tj68bpmk2a: ResNet Class
%\end{table}
\vspace{-0.4in}
\end{wrapfigure}
As mentioned earlier, learning learning rates on data points can be interpreted as learning a weighting on them.
Some recent works \cite{jiang2017mentornet,ren2018learning,shu2019meta,wang2019optimizing} 
have used meta-learning based approaches to dynamically assign weights to instances. 
In general, these works \cite{jiang2017mentornet, shu2019meta, wang2019optimizing} train a secondary neural network to assign weights to data-points
throughout the course of training, or \cite{ren2018learning} perform an online approximation of weights for each sample in the mini-batch.
%For a detailed review, we refer readers to the related work section.
These frameworks are Markovian in nature, since weights estimated at each time step are independent of past predictions.
In contrast, our framework treats learning rates on data as learnable parameters, and benefits from past history of optimization. 
In Table \ref{tab:impact_of_learnt_history}, we establish the importance of retaining optimization history, by
evaluating our framework without the use of history. 
Specifically, after each time step, we update the model parameters using the updated value of instance and class learning rates. 
Post model parameter update, we reset the instance and class learning rates to their initial value of 1. 
As shown in Table \ref{tab:impact_of_learnt_history}, not reusing the history of learnt learning rates leads to a 
significant drop in performance across all settings and architectures on CIFAR100. Another place of comparison with this prior work is in 
noisy setting (see Section \ref{subsec:robust_learning}), where we outperform these methods by a significant margin.

\begin{figure}
\begin{minipage}{\textwidth}
\begin{minipage}[]{0.3\textwidth}
    \centering
    \resizebox{1.0\columnwidth}{!}{
    \begin{tabular}{ccc}
      \toprule
                Dynamic weight decay   & \xmark         &  \cmark           \\
      \midrule
                 SGD                   & 77.5 $\pm$ 0.0 & 78.0 $\pm$ 0.1    \\
      \midrule
                Class-level            & 77.7 $\pm$ 0.1 & 78.3 $\pm$ 0.2    \\
                Instance-level         & 78.2 $\pm$ 0.1 & 78.6 $\pm$ 0.2    \\
      \bottomrule
    \end{tabular}
    % ResNet without weight-decay:
    % Instance: ynuxcitva5, lr_inst_param_3.2_lr_wd_param_0, (78.2)
    % Class: y78qknhc45, lr_class_param_1e-2_lr_wd_param_0, (77.7)
    % Baseline + wd: wubyak6828, lr_class_param_0_lr_wd_param_1e1, (78.0)
    }
    %\captionof{table}{Learning dynamic weight-decay improves SGD, class and instance-level optimizer on CIFAR100 with ResNet18.}
    %\label{tab:learnt_weight_decay}
\end{minipage}
\hfill
\begin{minipage}[]{0.7\textwidth}
    \centering
    \includegraphics[width=0.45\linewidth]{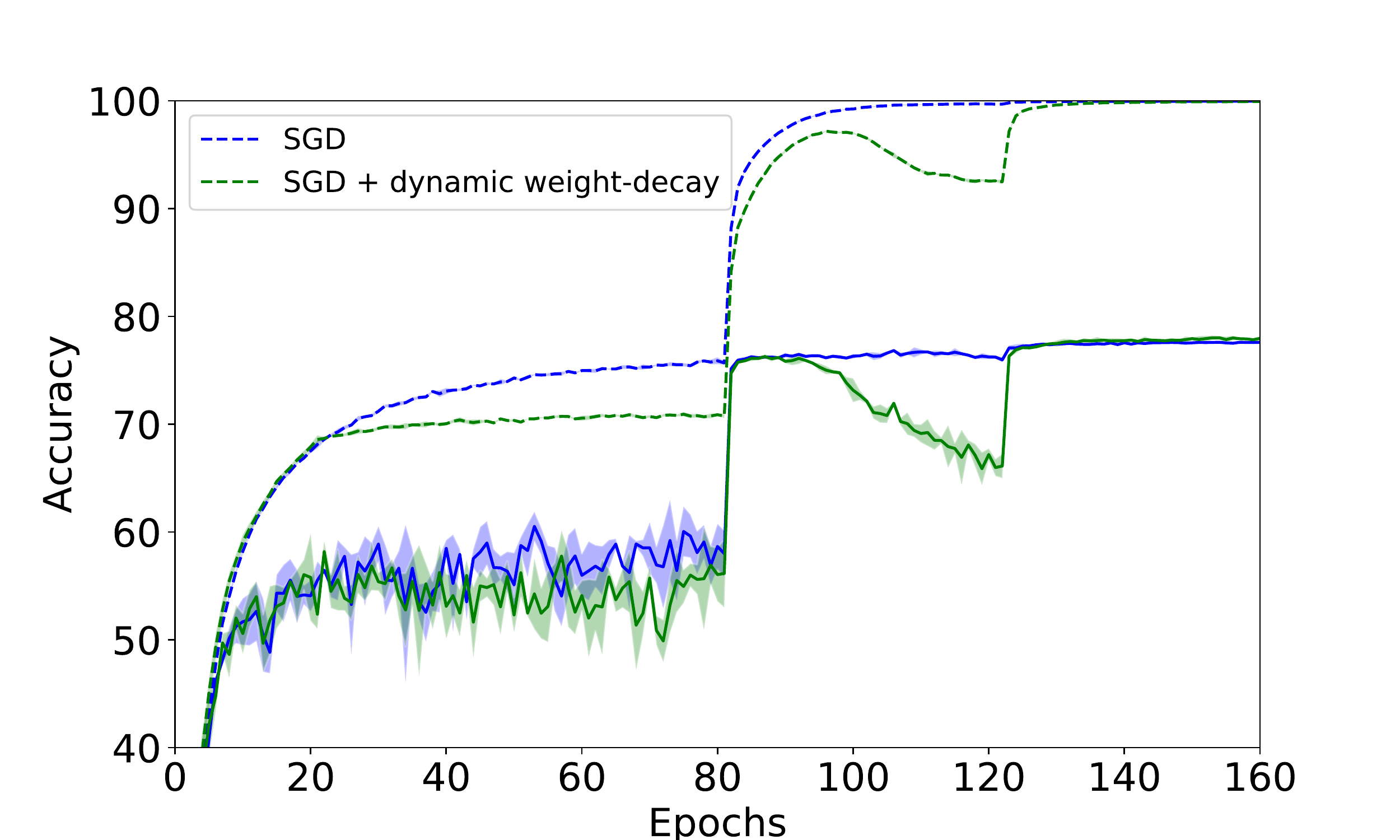}
    \includegraphics[width=0.45\linewidth]{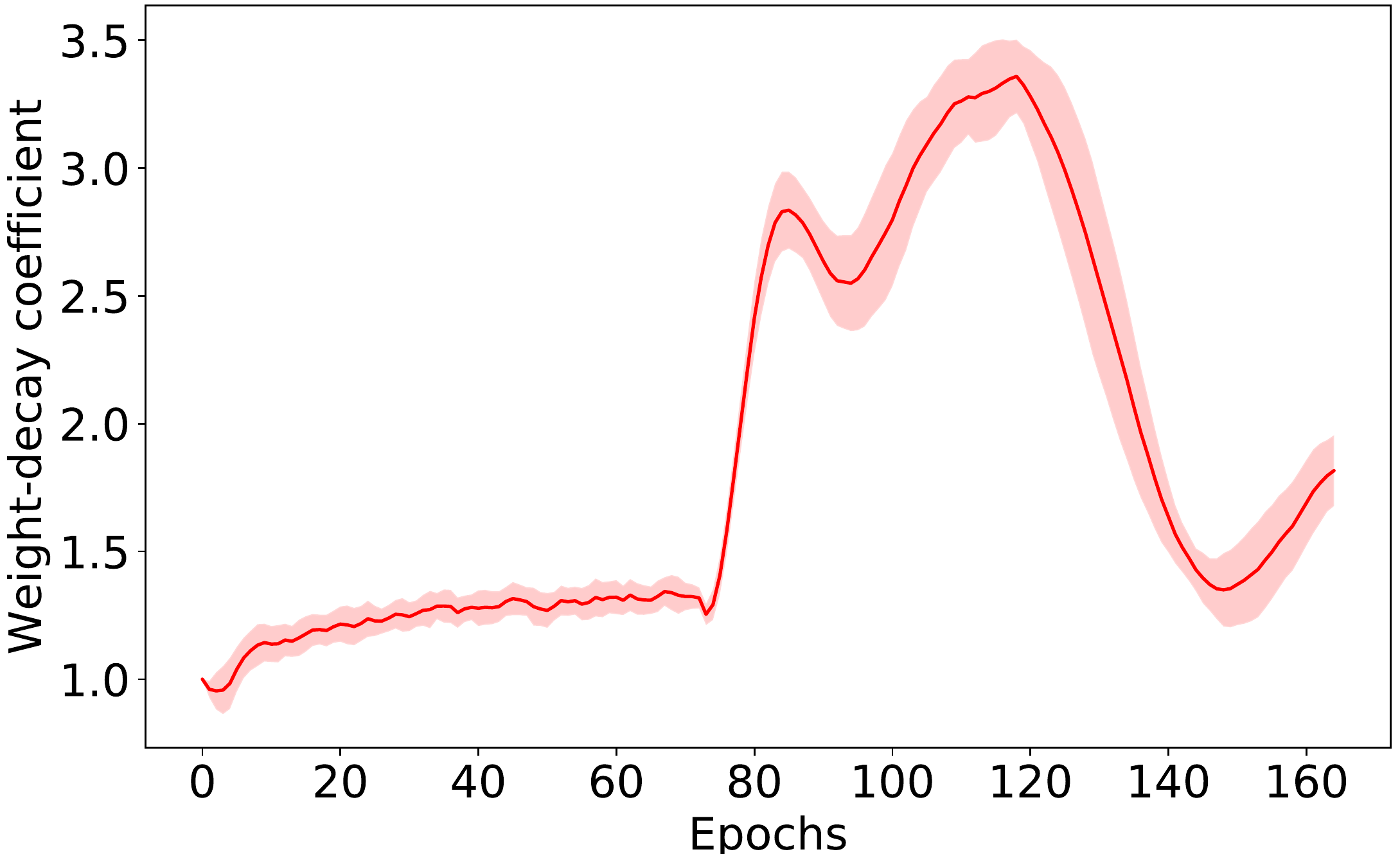}
    %\captionof{figure}{A table beside a figure}
    %\label{fig:learnt_weight_decay}
\end{minipage}
\end{minipage}
\caption{Evaluation of dynamic weight-decay on CIFAR100 with ResNet18.
  \textbf{Left}: Dynamic weight-decay improves all three optimizers. \textbf{Center}: Train (dashed) and validation (solid) accuracy for SGD optimizer, with
fixed and dynamic weight-decay. \textbf{Right}: Plot showing the value of dynamic weight-decay when learnt along with SGD. When the learning rate drops (epoch
80, 120), weight-decay coefficient increases so as to counteract overfitting, and obtains better performance at convergence.}
\label{fig:learnt_weight_decay}
\end{figure}

%\paragraph{Impact of learning non-stationary weight-decay regularization on optimization}
\paragraph{Importance of learning a dynamic weight-decay}
Recently, the importance of weight-decay in optimization of DNNs has gained much interest 
\cite{golatkar2019time, hoffer2018norm, loshchilov2018decoupled, zhang2018three}. 
\cite{golatkar2019time} empirically demonstrate that weight-decay plays an important role in the first few epochs, 
and does not play as much a role in the later stages of training. Our results indicates otherwise.
In this work, we show that learning a dynamic weight-decay leads to significant change
in SGD dynamics and also facilitates the learning of learning rates on data.
%In our work, our results show that learning a dynamic weight-decay
%leads to significant change in learning dynamics throughout the training process.
Figure \ref{fig:learnt_weight_decay} highlights the change in dynamics of SGD optimizer when weight-decay is learnt along with the model
parameters. Compared to baseline (fixed weight-decay), the learnt weight-decay adapts to different stages of optimization (see Figure
\ref{fig:learnt_weight_decay}, right).
More specifically, post learning rate drop, when model is most prone to overfitting, dynamic weight decay coefficient increases.
This leads to a temporary drop in performance, but results in better generalization at convergence. 
As see in the table (Figure \ref{fig:learnt_weight_decay}, left), learning a dynamic weight-decay improves performance of all three optimizers.

\subsection{Results on Robust Learning}
\label{subsec:robust_learning}
Learning instance-level learning rates can be useful when some of the 
labels in the dataset are noisy, where the framework should decrease learning rates on corrupt instances.
In this section, we validate our framework in a controlled corrupted label setting.

\begin{figure}
\begin{minipage}{\textwidth}
\begin{minipage}[]{0.3\textwidth}
    \centering
\resizebox{1.0\textwidth}{!}{%
  \begin{tabular}{lcc}
    \toprule
                                               &      40\%            &    60\%        \\ 
    \midrule
    Baseline  \cite{shu2019meta}               & $51.1 \pm 0.4$       & $30.9 \pm 0.3$  \\ 
    Focal Loss \cite{lin2017focal}             & $51.2 \pm 0.5$       & $27.7 \pm 3.7$  \\ 
    Co-teaching \cite{han2018co}               & $46.20 \pm 0.15$     & $35.7 \pm 1.2$        \\ 
    \midrule
    \multicolumn{3}{c}{Using 1000 clean images}\\
    \midrule
    MentorNet   \cite{jiang2017mentornet}      & $61.4 \pm 4.0$       & $36.9 \pm 1.5$         \\ 
    L2RW        \cite{ren2018learning}         & $60.8 \pm 0.9$       & $48.2 \pm 0.3$        \\ 
    MWNet       \cite{shu2019meta}             & $67.7  \pm 0.3$      & $58.8 \pm 0.1$        \\ 
    Ours (instance-level)                      & $\bf{69.0  \pm 0.2}$ & $\bf{59.6 \pm 0.3}$         \\ 
%    \midrule
%    Baseline on clean data (oracle)            & R      &  R   \\
    \bottomrule
  \end{tabular}
% Bolt ID for 60% MW-Net setting: b9ytm4gmi3, launched with multiple seeds: 4kmezkwkvq
% Bolt ID for 40% MW-Net setting: 54fpe2qxjp
}
    \captionof*{table}{Setting A}
    \label{tab:mwnet-noisy}
\end{minipage}
\hfill
\begin{minipage}[]{0.33\textwidth}
    \centering
\resizebox{1.0\textwidth}{!}{%
  \begin{tabular}{lcc}
    \toprule
                                               &   40\%     & 80\%   \\
    \midrule
    Baseline  \cite{ren2018learning}           & $50.66 \pm 0.24$       &       $8.0$        \\
    MentorNet PD \cite{jiang2017mentornet}     &       $56.9$           &      $14.0$              \\
    Data Parameters \cite{saxena2019data}      & $\bf{70.93 \pm 0.15}$  &  $35.8 \pm 1.0$  \\
    \midrule
    \multicolumn{3}{c}{Using 1000 clean images}\\
    \midrule
    MentorNet DD \cite{jiang2017mentornet}     &         $67.5$         &    $35.0$          \\
    L2RW         \cite{ren2018learning}        & $61.34 \pm 2.06$       &      -              \\
    Ours (instance-level)                      & $70.8 \pm 0.1$         &    $\bf{42.5 \pm 0.3}$                \\
 %   \midrule
%    Baseline on clean data (oracle)            & $74.18 \pm 0.19$       &  $58.32 \pm 0.53$   \\
    \bottomrule
  \end{tabular}
% Bolt ID for replication baseline on C100 with WRN using momentum (100% data): fr56e87pye (50.3 \pm 0.5)
% Bolt ID for replication baseline on C100 with WRN using momentum (98% data, 2% held-out for meta-training): tadcnf7shk (50.1 \pm 0.8)
% Bolt ID for our 40% result: 70.8 is obtained with momentum (at lr=0.1) dn945yzqpe; 
% Bolt ID for our 80% result:  42.5 is obtained without momentum (at lr=6.4) 4fsk4c7s86
}
  \captionof*{table}{Setting B}
\end{minipage}
\hfill
\begin{minipage}[]{0.33\textwidth}
\centering
\includegraphics[width=\linewidth]{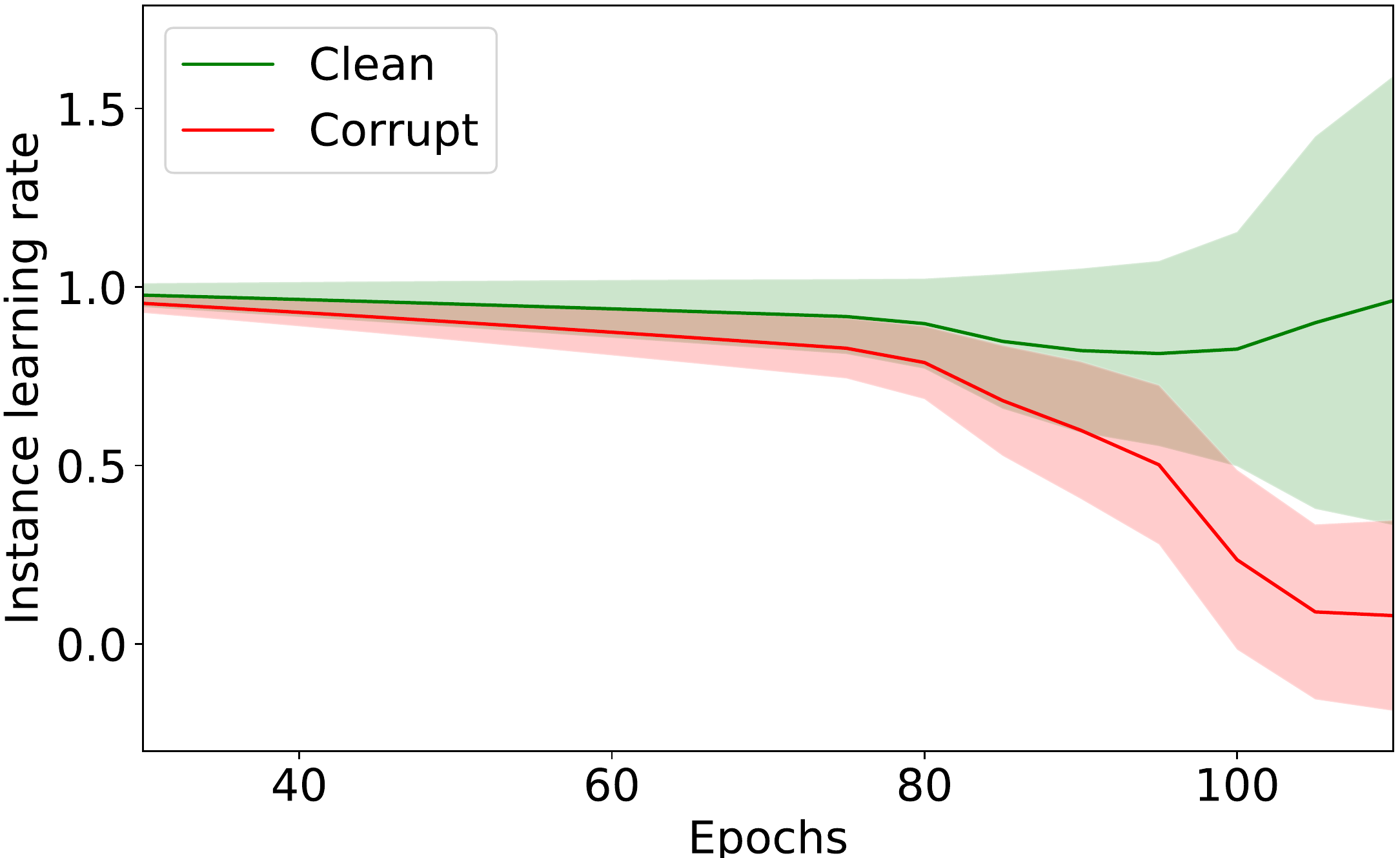}
%\caption{Mean and standard deviation of instance level learning rates.}
\end{minipage}
\end{minipage}
\caption{\textbf{Tables}: On CIFAR100, our methods outperforms state-of-the-art under varying levels of uniform noise in two settings (see text for details). 
%In both settings, we obtain competitive performance to recent state-of-the-art methods. 
\textbf{Figure}: Plot of mean and standard-deviation over instance-level
learning rates for clean and corrupts instances. Our framework learns to reduce the learning rate on corrupt instances.}
\label{fig:data_param_noisy}
\vspace{-0.3in}
\end{figure}

To compare with the relevant state-of-the-art, we follow the common setting in (\cite{jiang2017mentornet, ren2018learning}) to train deep CNNs, 
where the label of each image is independently changed to a uniform random class with probability $p$, where $p$ is noise fraction. 
The labels of validation data remain clean for evaluation. 
We compare our approach with recent state-of-the-art approaches in this setting \cite{jiang2017mentornet, ren2018learning, saxena2019data, shu2019meta}.
MentorNet \cite{jiang2017mentornet} and Meta-Weight-Net \cite{shu2019meta} train an auxillary neural network to assign weights to samples in the mini-batch.
L2RW \cite{ren2018learning} uses a held-out set to perform an online approximation of weights for samples in the mini-batch.
Data parameters \cite{saxena2019data} introduced learnable temperature parameters per data-point, which scale the gradient contribution of each data point. 
Unfortunately, all of these works report results in two distinct settings: setting A \cite{ren2018learning} and setting B \cite{shu2019meta}. 
While both settings use WRN-28-10, they differ in learning rate schedules (see details in supplementary). 
To make a fair comparion, we report results in both settings. 
Similar to setup in \cite{ren2018learning, shu2019meta}, we keep 1000 clean images as meta-set. As seen in Figure \ref{fig:data_param_noisy}, our method
outperforms other state-of-the-art methods in both settings under different levels of noise. 

\label{fig:data-param-noisy}

\subsection{Personalizing DNN models}
In a traditional setting, machine learning models are trained under empirical risk minimization (ERM) framework, where train and test set are assumed to be sampled
from the same distribution. These models are optimized to work well across the entire data distribution. However, this training setup would not be ideal when at
test time only a subset of train distribution is of interest. This situation can come up in various practical problems: (1) targeting a certain demographic for
recommender systems \cite{naumov2019deep}, (2) personalizing models in health for a certain anomaly or demographic \cite{jaques2017predicting, mikkelsen2018personalizing}, etc. 
In this setting, an important question one needs to answer is: \textit{What training data should I train the model on?}

%In this section, we consider the setting where the test distribution is a biased subset of train distribution, and the
%objective of optimization framework is to obtain a model which would do well on this biased distribution at inference. 
We simulate this scenario using the CIFAR100 dataset, which contains 100 fine grained classes, and 20 super classes (mutually disjoint, contains 5 classes).
In this scenario, despite having the entire dataset annotated, we are interested in one super class at test time.
Below we detail different methods which can be used in this scenario along with our proposed solution.

\textbf{Biased training}: The problem can be reduced to ERM framework by training the model on instances belonging to classes present in the super class. 
This approach makes an assumption that the other classes present in the dataset are completely disjoint from the super class. 
However, some of the discarded classes might share common low-level features which might be 
useful to train the early layers of deep neural network.

\textbf{Full training}: To address the aforementioned limitation, and take advantage of all the annotated data, one can train the model on all classes of the
CIFAR100 dataset. However, this approach makes an assumption that training on all classes would be beneficial for the classes present in super class.

\textbf{Transfer Learning}: Train model on all 100 classes (full training), followed by training on the classes present in super class (biased training). 
A limitation of this approach is that pretraining model on all 100 classes might bias the model.

\textbf{Our solution}: The limitation of approaches mentioned above lies in the fact that it involves making a hard choice regarding the classes present in the
training set. We relax this constraint by using dynamic class-level learning rates, which can guide the optimization process towards a biased subset
dynamically. The meta set is comprised of instances belonging to the super-class. 

We benchmark our method and the baselines in table below on the CIFAR100 dataset.
As seen in table in Figure \ref{fig:personalization_results}, using our proposed solution we outperform the other baselines by a significant margin.

\newcommand{\row}[1]{\multirow{2}{*}{{#1}}}

\begin{figure}
\hfill
\begin{minipage}{\textwidth}
    \begin{minipage}[]{0.6\textwidth}
    \centering
    \resizebox{1.0\columnwidth}{!}{%
      \begin{tabular}{lccccc}
        \toprule
        \row{Super Class}            &   \row{People}    & Aquatic               & \row{Vehicle 1}        & Electrical      &  \row{Reptiles}   \\
                                     &                   &  Mammals              &                        &  Devices        &                    \\
    %                                &     14            &   0                   &  18                    &    5            &    15     \\
        \midrule
        Biased Training              &   $54.1 \pm 1.3$  & $69.7 \pm 4.5$   &  $91.1 \pm 0.5$        & $87.0 \pm 0.5$  &   $77.0 \pm 0.5$    \\
        Full Training                &   $55.5 \pm 1.6$  & $61.1 \pm 1.4$        &  $84.7 \pm 0.5$        & $77.5 \pm 1.1$  &   $65.1 \pm 1.4$            \\
        Transfer Learning            &   $54.6 \pm 1.4$  & $66.4 \pm 4.1$        &  $91.0 \pm 0.2$        & $87.7 \pm 0.7$  &   $76.2 \pm 0.3$     \\
        \midrule
        Ours (class-level)         &   $\bf{61.0 \pm 1.2}$     & $\bf{70.7 \pm 0.3}$                &  $\bf{93.0 \pm 0.5}$   & $\bf{88.1 \pm 0.8}$     &   $\bf{79.5 \pm 0.7}$        \\
        \bottomrule
      \end{tabular}
    }

\end{minipage}
\hfill
\begin{minipage}[]{0.3\textwidth}
    \includegraphics[width=\linewidth]{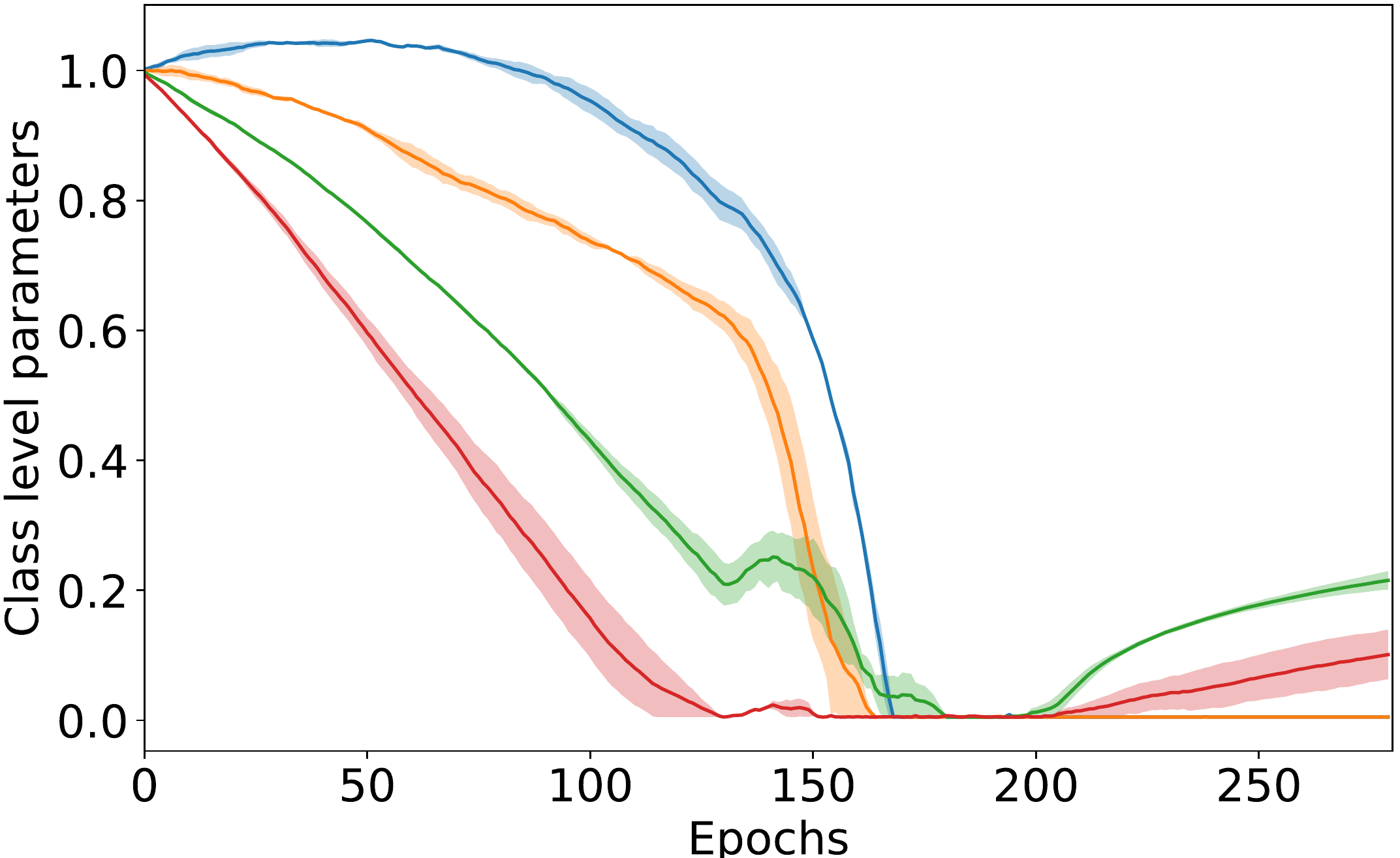}
\end{minipage}
\hfill
\end{minipage}
\caption{\textbf{Table:} For the task of personalization towards a super-class, using dynamic class level learning rates outperforms baselines which 
involve a static choice of train data. \textbf{Figure:} 
Repetable trajectories of learning rate for classes not present in the super class.}

\label{fig:personalization_results}
\end{figure}

\section{Related Work}
Optimization of DNNs has gained a lot of interest recently \cite{andrychowicz2016learning, hoffer2018norm, johnson2018training, kidambi2018insufficiency,
loshchilov2018decoupled, martens2015optimizing, zhang2019lookahead} . While a full detailed review is beyond the scope of this paper, here, 
we give a brief overview of related work most relevant to the material present in the paper. 

Learning adaptive learning rates on data can be interpreted as learning a weighting on each data point. 
\cite{jiang2017mentornet, shu2019meta, wang2019optimizing} train an auxillary neural network to assign weights to data points.
\cite{ren2018learning} performs an online approximation, where it uses one step look ahead on meta-set to estimate the weights for samples in the mini-batch.
These approaches are Markovian in nature, since weights estimated at each time step are independent of past-predictions.
In contrast, our method leverages the past history of optimization, and outperforms these state-of-the-art methods for 
robust learning (see Section \ref{subsec:robust_learning}).

Our work also has connections to importance sampling. 
\cite{jiang2019accelerating, johnson2018training, katharopoulos2018not} propose approximations for the gradient norm of instances which are used for sampling data points.
Theoretically, sampling data points with high gradient norm should lead to faster convergence. However, this would not work well in real world dataset which
contain noisy data. In the same spirit, our work also has connections to the field of curriculum
learning \cite{bengio2009curriculum, hacohen2019curriculum, spitkovsky2009baby} and self-paced learning \cite{fan2017self, lee2011learning, pi2016self, supancic2013self}. 
These approaches either design hand-crafted heuristics, or use loss value of data points as a proxy to decide the ordering of data.
All these approaches require coming up with a heuristics which might not work from one problem domain to another. In contrast, 
our work can be interpreted as a soft differentiable form of importance sampling, where the importance of a sample (learning rate) or curriculum is 
learnt through meta gradient.

Data parameters \cite{saxena2019data} introduced learnable temperature parameters for each instance and class in the dataset.
These parameters controlled the gradient contribution for each data point, and were learnt using gradient descent.
In similar spirit, \cite{barron2019general} introduced learnable robustness parameter per data point, which generalized different regression losses. 
Both of these works learn parameters per data point for robust estimation in classification or regression. In comparison, our formulation is more 
generic and can admit any differentiable loss function. More importantly, due to its meta-learning framework, our approach allows for robust estimation of noise as
well as personalization.

Other work has proposed optimizing a few hyperparameters, such as kernel parameters \cite{chapelle2002choosing}, weight decay \cite{bengio2000gradient} or
others \cite{maclaurin2015gradient}, using gradients during training. However, to the best of our knowledge, 
none have done so for learning dynamic learning rates across training per se, nor done so at scale 
(e.g. one rate per instance) to achieve state-of-the-art performance.   

%Learning with parameter per instance
%Data parameter paper
%CVPR paper

\section{Conclusion}
In this paper, we have proposed an optimization framework
which accounts for differences in loss function characteristics across instances and classes present in the dataset.
More specifically, our framework learns a dynamic learning rate for each instance and class present in the dataset. 
Learning a dynamic learning rate allows our framework to focus on different modes of training data. 
For instance, when presented with noisy dataset, our framework reduces the learning rate on noisy instances, 
and focuses on optimizing model parameters using clean instances. 
When applied for the task of image-classification, across different CNN architectures, our framework outperforms standard optimizers. 
Finally, for the task of personalization of machine learning models towards a known data distribution, our framework outperforms 
strong baselines.

\newpage
{\small
\bibliographystyle{ieee_fullname}
\bibliography{egbib}
}

%\newpage
%\input{./tex_files/supplementary.tex}
\end{document}